\documentclass[letterpaper, 10 pt, conference]{ieeeconf}
\IEEEoverridecommandlockouts                              

\usepackage[utf8]{inputenc}

\usepackage[x11names]{xcolor}
\usepackage{dsfont} 
\usepackage{bm}
\usepackage{mathtools} 
\usepackage{graphicx}%
\usepackage{multirow}%
\usepackage{amsmath,amssymb,amsfonts}%
\usepackage{amsthm}%
\usepackage{mathrsfs}%
\usepackage{xcolor}%
\usepackage{textcomp}%
\usepackage{manyfoot}%
\usepackage{booktabs}%
\usepackage{algorithm}%
\usepackage{algorithmicx}%
\usepackage{algpseudocode}%
\usepackage{listings}%

\usepackage{tikz}
\usetikzlibrary{angles,quotes}

\newtheorem{assumption}{Assumption}

\usepackage{hyperref}
\hypersetup{
    colorlinks=true,
    linkcolor=black,
    filecolor=black,
    urlcolor=blue,
    citecolor=blue,
    }
\usepackage{todonotes}

\usepackage{blindtext}

\usepackage{booktabs}

\title{\LARGE \bf Learning a Tracking Controller for Rolling $\mu$bots}

\author{Logan E. Beaver$^{1}$, Max Sokolich$^{2}$. Suhail Alsalehi$^{1}$, Ron Weiss$^{3}$, Sambeeta Das$^{2}$, Calin Belta$^{1}$
\thanks{*This work was supported by 
the National Science Foundation under grant GCR 2219101 and by the National Institute of Health under grant R35GM147451.}
\thanks{$^{1}$Division of Systems Engineering, Boston University, Boston, MA 02215, USA
        {\tt\small \{lebeaver,alsalehi,cbelta\}@bu.edu}}%
\thanks{$^{2}$Department of Mechanical Engineering, University of Delaware, Newark, DE 29716, USA
        {\tt\small \{sokolich,samdas\}@udel.org}}%
\thanks{$^{2}$Department of Biological Engineering, Massachusetts Institute of Technology, Cambridge, MA 02142, USA
        {\tt\small rweiss@mit.edu}}%
}

\begin{document}

\maketitle
\thispagestyle{empty}
\pagestyle{empty}
\begin{abstract}
Micron-scale robots ($\mu$bots) have recently shown great promise for emerging medical applications. Accurate controlling $\mu$bots, while critical to their successful deployment, is challenging. 
In this work, we consider the problem of tracking a reference trajectory using a $\mu$bot in the presence of disturbances and uncertainty.
The disturbances primarily come from Brownian motion and other environmental phenomena, while the uncertainty originates from errors in the model parameters.
We model the $\mu$bot as an uncertain unicycle that is controlled by a global magnetic field. To compensate for disturbances and uncertainties, we develop a nonlinear mismatch controller.
We define the \emph{model mismatch error} as the difference between our model's predicted velocity and the actual velocity of the $\mu$bot.
We employ a Gaussian Process to learn the model mismatch error as a function of the applied control input.
Then we use a least-squares minimization to select a control action that minimizes the difference between the actual velocity of the $\mu$bot and a reference velocity.
We demonstrate the online performance of our joint learning and control algorithm in simulation, where our approach accurately learns the model mismatch and improves tracking performance. 
We also validate our approach in an experiment and show that certain error metrics are reduced by up to $40\%$.
\end{abstract}

\section{Introduction} \label{sec:intro}

Interest in micron-scale robots ($\mu$bots) has grown exponentially in recent decades \cite{honda1996micro}.
Medical applications have been of particular interest, including drug delivery \cite{sitti2015biomedical, troccaz2008development}, biopsy \cite{Barcena2009ApplicationsBiomedicine}, microsurgery \cite{Guo2007MechanismApplication}, and cellular manipulation \cite{Sakar2011WirelessMicrotransporters,Jagerpaper,kim2013fabrication,Steager2013AutomatedMicrorobots}.
Despite these advances, there are numerous challenges associated with the control of $\mu$bots.
The extremely small scale of $\mu$bots incentivizes novel actuation techniques, such as electrophoretic \cite{kim2015electric}, optical \cite{palima2013gearing}, magnetic \cite{chowdhury2016towards}, thermal  \cite{ErdemThermallyMicrorobot}, or by attachment to swimming microorganisms \cite{Behkam2007BacteriaMicrorobots}.

Our $\mu$bot is controlled by a rotating 3D magnetic field.
The field induces a rotating moment on the $\mu$bot, which causes it to roll along the substrate surface during experiments.
As a consequence, this method uses significantly less energy than other actuation methods, e.g., translating particles using strong magnetic gradients.
A similar control technique has been previously used to control the micron scale ``rod-bot'' \cite{pieters2015rodbot} for micron-scale manipulation.
The $\mu$bot we control is spherical, non-toxic to living cells, and can be embedded within cells without damaging them \cite{Rivas2022}.
This makes it an ideal candidate for emerging medical applications involving cellular manipulation.
However, the small size of the $\mu$bot also implies that Brownian motion plays a significant role in its dynamics (see \cite{das2018experiments,vinagre2016there}), and modeling error makes the $\mu$bot difficult to control accurately.
The readers are referred to \cite{xu2015magnetic} for further details on different actuation techniques and motion control strategies for robots at the micron scale, .

In this article, we develop a joint learning and control approach to improve the tracking capabilities of rolling $\mu$bots.
A related vision-based control system to manipulate rolling $\mu$bots was presented in \cite{tang2022vision}, where the authors used closed-loop visual feedback to navigate through an environment with impurities and obstacles. 
In contrast, we propose an open-loop strategy that takes the desired velocity as an input and yields a corrected control signal that minimizes the difference between the desired and actual velocities of the $\mu$bot.
As a consequence, our approach is straightforward to include as an integral step for receding horizon control and other closed-loop feedback strategies. 
Compared to related learning-based approaches, e.g., controlling swimming micron-scale robots \cite{behrens2022smart}, our approach is model-based and explicitly embeds the learning within the controller.

Inspired by \cite{Greeff2021ExploitingProcesses}, in this work we derive a controller to minimize the $\emph{nonlinear mismatch error}$, that is, the error between our nonlinear $\mu$bot model and the actual dynamics.
We achieve this in three steps. First, we invert an empirically derived $\mu$bot model to convert a desired velocity into a desired control action.
Then, we use the learned nonlinear mismatch error and least-squares optimization to generate a corrected control signal.
The corrected control signal exploits the learned error to minimize the difference between the desired and actual velocity of the $\mu$bot.
In comparison, \cite{Greeff2021ExploitingProcesses} updates the desired velocity of the system before inverting the dynamics. 
The contributions of this article are as follows: 
\begin{itemize} \itemsep0em
    \item we extend the \emph{inverse nonlinear mismatch} approach of \cite{Greeff2021ExploitingProcesses} from a 1D regression problem with linear dynamics to a $2$D trajectory tracking problem;
    \item we derive an explicit functional form of the $\mu$bot's input-output velocity error to demonstrate that model parameter fitting is insufficient for accurate control - this also motivates the development of nonlinear control techniques; 
    \item we derive an novel control strategy to correct the nonlinear model mismatch error by explicitly embedding a Gaussian Process regression model within a least-squares optimization problem; and 
    \item we demonstrate improvement in the $\mu$bot's tracking capability in simulation and experiment, and we show that our online learning approach is real-time implementable.
\end{itemize}

The remainder of this article is organized as follows: We present our experimental setup in Section \ref{sec:experimentSetup} and formulate the tracking problem in Section \ref{sec:problem}.
We present our learning approach in Section \ref{sec:learning}.
Simulation and experimental results are included in Section \ref{sec:results}, and we draw conclusions and discuss future work in Section \ref{sec:conclusions}.

\section{Experimental Setup} \label{sec:experimentSetup}
 
\begin{figure*}[ht]
    \centering
    \vspace{1em}
    \includegraphics[width=12cm]{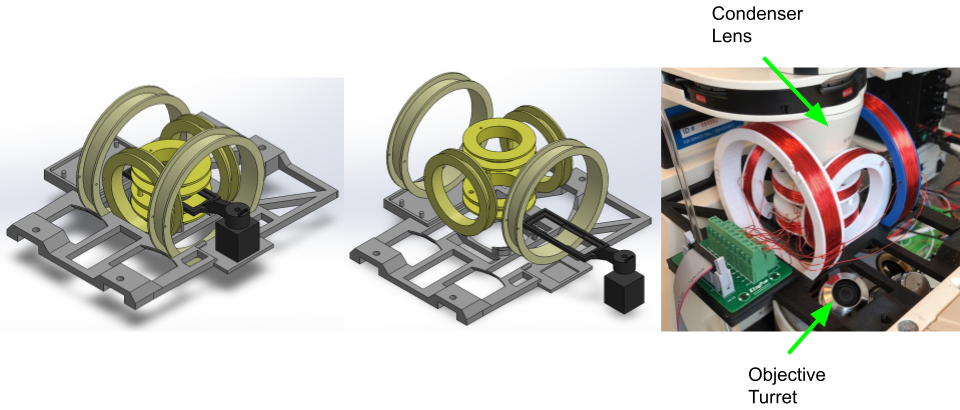}
    \caption{Design schematics and image of the 3D Helmholtz-based System}
    \label{fig:Helmholtz}
\end{figure*}

In order to generate the magnetic fields necessary to actuate the rolling $\mu$bots, 6 Helmholtz coils are designed and arranged in parallel pairs as shown in Fig. \ref{fig:Helmholtz}. The coils are mounted on a Zeiss Axiovert 100 inverted microscope. To power the coils, we use an Arduino Mega micro-controller connected to a Jetson Xavier NX single board computer similar to a Raspberry Pi. The Jetson Xavier NX is capable of running a full Linux distribution with the help of a keyboard, mouse and monitor. A custom tracking and control program is written in python to read incoming images from a FLIR BFS-U3-28S5M-C USB 3.1 Blackfly® S Monochrome Camera.  The continuous stream of images is analyzed in Python's OpenCV library, which is used to extract position and velocity data for detected microrobots. 

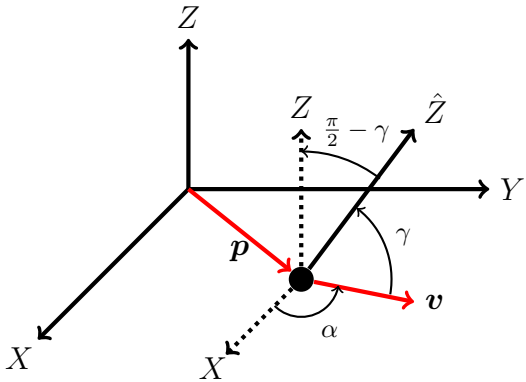
\begin{figure}[ht]
    \centering
    \begin{tikzpicture}
    \draw[->, ultra thick] (0,0) -- (4,0) node[right] {\large$Y$};
    \draw[->, ultra thick] (0,0) -- (0,2) node[above] {\large$Z$};
    \draw[->, ultra thick] (0,0) -- (-2,-2);
    \node at (-2.25,-2.25) {\large$X$};    \node[circle,fill=black,minimum size=3pt]
    (bot) at (1.5,-1.2) {};
    \draw[->, ultra thick,color=red]
    (0,0) -- node[below,color=black] {\large$\bm{p}$}  (bot);
    \draw[->, ultra thick,color=red]
    (bot) --  (3,-1.5) node[color=black,right] (fwd) {\large$\bm{v}$};
    \draw[->, ultra thick,dotted]
    (bot) -- ++(-1,-1) node[shift={(-5pt, -5pt)}] (X) {\large${X}$};
    \pic [draw, ->, "$\LARGE\alpha$", angle eccentricity=2.5, thick,pic text options={shift={(0pt,0.5cm)}}] 
    {angle = X--bot--fwd};
    \draw[->, ultra thick] (bot) -- (3,0.8) node[shift={(8pt,8pt)}] (attd) {\large$\hat{Z}$};
    \pic [draw, ->, angle radius = 1.2cm, "$\LARGE\gamma$", angle eccentricity=0.5, thick,pic text options={shift={(.8cm,10pt)}}] 
    {angle = fwd--bot--attd};
    \draw[->, ultra thick,dotted] (bot) -- ++(0,2) node[above] (z2) {\large$Z$};
    \pic [draw, ->, angle radius=1.7cm, "$\LARGE\frac{\pi}{2}-\gamma$", angle eccentricity=1.0, thick,pic text options={shift={(5pt,10pt)}}] 
    {angle = attd--bot--z2};
    \end{tikzpicture}
    \caption{Schematic illustrating the notation. The solid axes $X$, $Y$, $Z$ define the Cartesian coordinate system. The motion of the $\mu$bot is the in the $(X,Y)$ - plane. The fixed attitude angle $\gamma$ is out of the plane, while the heading angle $\alpha$ is in plane.  The frequency $f$ determines the $\mu$bot's forward speed along $\bm{v}$, which is in the $(X,Y)$ - plane.}  
    %
    \label{fig:Rot}
\end{figure}

Action commands in the form of a heading angle $\alpha$ to steer the $\mu$bot, a constant attitude angle $\gamma$, and a frequency $f$ to set the speed at which the magnetic field rotates are sent to the Arduino over a serial communication protocol.  These action/input commands are converted to a 3D rotating magnetic field. 
Because current is proportional to the magnetic field generated from electromagnets, the magnetic field $\bm{B} = [B_x,B_y,B_z]^T$ is mapped to the heading, attitude, and frequency signals via
\begin{equation} \label{eq:magnet}
    \bm{B} =
    \begin{bmatrix}
        \cos(\gamma)\cos(\alpha)\cos(2\pi f t) + \sin(\alpha)\sin(2\pi f t) \\
        -\cos(\gamma)\sin(\alpha)\cos(2\pi f t) + \cos(\alpha)\sin(2\pi f t) \\
        \sin(\gamma)\cos(2\pi f t)
    \end{bmatrix}, 
\end{equation}
where $\gamma\in[-\pi, \pi]$ is a fixed attitude angle, $\alpha\in[-\pi, \pi]$ is the heading angle, 
and $f\in\mathds{R}$ is the rolling frequency.
The angles and coordinate system are depicted in Fig. \ref{fig:Rot}.
Note that since $\bm{B}$ is 3 Dimensional, half of the required current is sent to each pair of parallel electromagnets to achieve the desired magnetic field strength.

The $\mu$bots are constructed by plasma cleaning a plain glass slide on high for 5 minutes, wherein 24 $\mu m$ paramagnetic, fluorescent microspheres (Spherotech® FCM-10052-2) mixed with ethanol are drop casted and left to dry.
The microspheres are coated with a $100$ nm thick layer of Nickel in a dual electron beam deposition chamber, which increases the $\mu$bot's magnetic moment.
Due to the inherent surface properties of the $\mu$bot and the substrate surface, there are often very large attractive forces that result in the $\mu$bot sticking to the surface, hindering its motion.  This is highly unpredictable and quite common despite adequate cleaning of the microscope slide surface. As a result, two additions were made to the experimental procedure to help reduce the likelihood of sticking. Firstly, the plasma cleaned glass slide was additionally incubated in a PFOTS (1H,1H,2H,2H-perfluorooctyltrichlorosilane) vapor at 85°C for 30 minutes.  This results in a hydrophobic surface that allows the $\mu$bot to more easily roll across the surface. 
Secondly, instead of suspending the $\mu$bot's in DI water, they are suspended in a 0.1 $\%$ solution of Sodium Dodecyl Sulfate, which is a surfactant.  Although this reduces the rolling speed of the microrobot due to the increased viscosity, it significantly reduces the chances of the microrobot sticking.

\section{Problem Formulation and Approach} \label{sec:problem}

The $\mu$bot is a roughly spherical magnetic particle that we control using a 3D magnetic field.
We control the $\mu$bot by continuously rotating the magnetic field using \eqref{eq:magnet}, which induces a rotational moment in the $\mu$bot and causes it to roll along the substrate surface.
Varying the frequency $f$ affects the forward speed of the $\mu$bot, while varying the heading angle $\alpha$ affects its heading direction; these are depicted in Fig. \ref{fig:Rot}.
Based on the rolling motion of the $\mu$bot, we model it as a unicycle subject to a generalized disturbance term \cite{Yang2020}:
\begin{equation} \label{eq:dynamics}
    \begin{aligned}
        \dot{\bm{p}} = a_0 f
        \begin{bmatrix}
        \cos(\alpha) \\ \sin(\alpha)    
        \end{bmatrix}+ \bm{D},
    \end{aligned}
\end{equation}
where $\bm{p} \in \mathds{R}^2$ is the position in a given reference frame, $a_0\in\mathds{R}_{>0}$ is an empirically determined effective radius of the $\mu$bot, and $f, \alpha$ are the $\mu$bot's rotation frequency and heading angle, respectively.
Finally, $\bm{D}\in\mathds{R}^2$ is a disturbance term that captures Brownian motion and other micron-scale disturbances.
Note that we subsequently justify that 1) the control actions are identical to the heading angle and rolling frequency of \eqref{eq:magnet}, and 2) the frequency $f$ can be fixed in practice, and thus we consider only a single control input $\alpha$.

Our objective is to follow a reference trajectory. Let 
$\bm{v}=\dot{\bm{p}}$ denote the velocity of the $\mu$bot (see Fig. \ref{fig:Rot}). Given a desired velocity signal $\bm{v}^d(t)$, we seek to find the optimal control input $\alpha$ such that the difference between $\dot{\bm{p}}(\alpha(t))$ and $\bm{v}^d(t)$ is minimized. This control achieves our  objective given that the $\mu$bot starts on the reference trajectory. This approach is useful for high-level planners, e.g., those using RRT* and MPC, as they can generate trajectories using only the kinematic model $\dot{\bm{p}}\approx\bm{v}^d$.
This decouples trajectory tracking from high-level motion planning, which is an area of ongoing research.
To achieve this, we present our working assumptions for our tracking controller next.

\begin{assumption} \label{smp:disturbance}
    The environmental disturbance $\bm{D}$ and any error in our model of the true dynamics \eqref{eq:dynamics} are isotropic, i.e., they do not depend on the $\mu$bot's position $\bm{p}$.
\end{assumption}

\begin{assumption} \label{smp:alignment}
    There error in aligning the $\mu$bot with the magnetic field is negligible, i.e., $\alpha$ and $f$ in \eqref{eq:magnet} and \eqref{eq:dynamics} are identical.
\end{assumption}

Assumptions \ref{smp:disturbance} and \ref{smp:alignment} simplify the learning process and are reasonable for the laboratory environment.
The disturbance affecting the $\mu$bot is primarily Brownian motion, which acts uniformly at random to disturb the velocity.
Assumption \ref{smp:disturbance} could be relaxed by having the $\mu$bots infer hydrodynamic disturbances caused by heat, density, and chemical concentration differences, e.g., using an approach similar to \cite{Dongsik2017MotionVehicles}.
In previous work, we have also found that the alignment of $\mu$bots to the global magnetic field is nearly instantaneous (see \cite{Beaver2021AMicrorobots}), which justifies Assumption \ref{smp:alignment}.

\begin{assumption} \label{smp:constantSpeed}
    The $\mu$bot is controlled to roll at a fixed rate, i.e., $f(t)$ is a known constant selected a priori.
\end{assumption}

Assumption \ref{smp:constantSpeed} does not affect the derivation of our controller, as we only employ it when training our machine learning algorithm and solving the least-squres optimization.
Relaxing this assumption increases the amount of training data and learning time, but not prohibitively so.
Furthermore, operating $\mu$bots with a constant rolling frequency is common practice \cite{Rivas2022}.

Our technical approach is as follows:
First, we parameterize the $\mu$bot's dynamics with an approximate model, and we derive the functional form of the model error.
We show that the domain of the model error function is a subset of the state space, which we use as the features (inputs) for a machine learning algorithm.
After learning the model error, we employ least-squares optimization to minimize the difference between the predicted and actual velocity of the $\mu$bot.
Note that we do not minimize the predicted velocity error directly.
Instead, we adjust the control signal sent to the $\mu$bot to compensate for the model error indirectly.

\section{Nonlinear Mismatch Controller} \label{sec:learning}

The unicycle model satisfies the property of \emph{differential flatness} with the output variable $\bm{p}$, that is, we can change the coordinates of our unicycle dynamics \eqref{eq:dynamics} to only consider the variable $\bm{p}$ and its derivative $\dot{\bm{p}}\coloneqq\bm{v}$ (see \cite{Sira-Ramirez2018DifferentiallySystems}).
Furthermore, while the frequency $f$ is fixed under Assumption \ref{smp:constantSpeed}, it is straightforward to relax this assumption for our analysis.
In fact, allowing a variable frequency $f$ only introduces computational complexity while training the machine learning model and solving the least-squares optimization problem.
The mapping for our rolling dynamics is
\begin{equation} 
    \begin{aligned} \label{eq:model-flatness}
    \alpha &= \arctan\Bigg({\frac{v_y - D_y}{v_x - D_x}}\Bigg), \\
    f &= \frac{||\bm{v} - \bm{D} ||}{a_0},
    \end{aligned}
\end{equation} 
where $\bm{D}=[D_x\;D_y]$ and $\bm{v}=[v_x\;v_y]$. 
In reality we do not know the actual value of $a_0$, nor do we know the stochastic disturbance $\bm{D}$.
Thus, we denote our approximate model parameters using $\hat{\cdot}$.
In particular, $\hat{a}_0$ is a constant scalar that estimates $a_0$ and $\hat{\bm{D}}$ is a constant vector that estimates the disturbance $\bm{D}$.
With these estimates, it is possible to convert a desired velocity $\bm{v}^d$ into a control signal $\alpha^d$ using \eqref{eq:model-flatness}, i.e.,
\begin{equation}
\begin{aligned} \label{eq:model-approx}
    \alpha^d &= \arctan\Bigg(\frac{v_y^d - \hat{D}_y}{v_x^d - \hat{D}_x}  \Bigg), \\
    f^d &= \frac{||\bm{v}^d - \hat{\bm{D}}||}{\hat{a}_0}.
\end{aligned}
\end{equation}
This enables us to convert the desired velocity $\bm{v}^d$, which is defined on a Cartesian basis, into control actions for the $\mu$bot.
However, due to the inherent inaccuracies of our model, na\"ively applying the control input $\alpha^d$ leads to tracking error.
Substituting the approximate model \eqref{eq:model-approx} into the dynamics \eqref{eq:dynamics} yields the actual velocity of the $\mu$bot in the form
\begin{equation} \label{eq:realizedSpeed}
    \bm{v} = \frac{a_0}{\hat{a}_0}\bm{v}^d + \bm{D} - \frac{a_0}{\hat{a}_0}\hat{\bm{D}}.
\end{equation}
Note that the actual velocity contains the nonlinear product of $\frac{a_0}{\hat{a}_0}$ and $\hat{\bm{D}}$.
This explains why a model parameter estimation alone is insufficient to achieve a desired trajectory, as the error in our model parameters is amplified by this nonlinearity.

To enhance our ability to track the desired trajectory beyond parameter estimation, we follow the approach outlined by \cite{Greeff2021ExploitingProcesses} and explicitly define a velocity error $\bm{v}^e$,
\begin{equation} \label{eq:vError}
    \bm{v}^e \coloneqq \bm{v} - \bm{v}^d,
\end{equation}
which, using \eqref{eq:realizedSpeed}, we can write in closed form as
\begin{equation} \label{eq:velocity-error-full}
    \bm{v}^e = \bm{v}^d\Big(\frac{a_0}{\hat{a}_0} - 1\Big) + \bm{D} - \frac{a_o}{\hat{a}_o}\hat{\bm{D}}.
\end{equation}
Note that \eqref{eq:model-approx} enables us to freely convert between $\bm{v}^d$ and $\alpha^d, f^d$.
Thus, the right hand side of \eqref{eq:velocity-error-full} is a function of $\alpha^d$, $f^d$, the domain of $\bm{D}$, and the domain of $\hat{\bm{D}}$.
This implies that, under Assumptions \ref{smp:disturbance}--\ref{smp:constantSpeed}, we can completely capture the behavior of the velocity error as some function $\bm{v}^e$ using machine learning with $\alpha^d$ as the only feature.
In particular, we approximate $\bm{v}^e$ using Gaussian Process (GP) regression.
We train the GP online using experimental data, where the Cartesian axes of $\bm{v}^e$ are each captured by a GP.
In particular, we use \eqref{eq:vError} to generate training data by applying a known sequence of control inputs; we discuss this process further in Section \ref{sec:howto-learn}.

\begin{figure*}[ht]
\vspace{1em}
    \centering
    \begin{tikzpicture}
   \draw [fill=yellow!40!white]
       (1.38,-1) -- (1.38,1.1) -- (6.7,1.1) -- (6.7,-1) -- cycle;
    \node[align=center,rounded corners,draw,minimum size=1.5cm] (A) at (0,0) {Reference\\ Trajectory};
    \node[fill=white,align=center,rounded corners,draw,minimum size=1.5cm] (B) at (2.75,0) {Inverted\\ Model \eqref{eq:model-flatness}};
    \node[fill=white,align=center,rounded corners,draw,minimum size=1.5cm] (C) at (5.25,0) {Nonlinear\\ Mismatch};
    \node[fill=white,align=center,rounded corners,draw,minimum size=1.5cm] (D) at (7.75,0) {Plant\\ ($\mu$bot)};
    \node[align=center,rounded corners,draw,minimum size=1.5cm] (E) at (10,0) {Sensor +\\ Observer};
	\draw[-stealth,ultra thick] (A) -- (B);
    \draw[-stealth,ultra thick] (A) -- (1.5,0) -- (1.5,1.025) -- (4.75,1.025) -- (4.75,0.75);
    \draw[-stealth,ultra thick] (B) -- (C);
    \draw[-stealth,ultra thick] (C) -- (D);
    \draw[-stealth,ultra thick] (D) -- (E);
    \draw[-stealth,ultra thick] (E) -- (11.7,0) -- (11.7,1.2) -- (5.25, 1.2) -- (5.25, 0.75);
    \draw[-stealth,ultra thick] (E) -- (11.7,0) -- (11.7,-1.2) -- (0, -1.2) -- (A);
    \node at (1.15, 0.25)  {$\bm{v}^d$};
    \node at (4, 0.25)  {$\alpha^d$};
    \node at (6.45, 0.25)  {$\alpha^*$};
    \node at (8.75, 0.25)  {$\bm{v}$};
    \node at (11.25, 0.25)  {$\bm{v}(t_k)$};
    \end{tikzpicture}
    \caption{A control block diagram showing how our proposed system (yellow box) transforms the desired velocity signal ($\bm{v}^d$) into a heading angle ($\alpha^*$) such that the difference between $\bm{v}^d$ and $\bm{v}$ is minimized.}
    \label{fig:control-diagram}
\end{figure*}
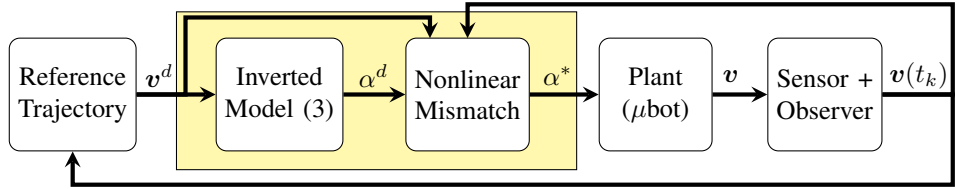

A GP is completely defined by its mean $\mu(\alpha)$ and kernel (or covariance) $K(\alpha,\alpha')$ functions for two features $\alpha, \alpha'$.
The prior of the mean is generally zero, while the kernel describes a statistical distribution over a function space.
For accurate regression, the kernel should be a basis for the underlying function $\bm{v}^e(\alpha$).
After learning the velocity error, the GP takes the heading angle $\alpha$ as an input and produces a Gaussian distribution over each component of $\bm{v}^e$.
The mean of this distribution predicts the expected value of $\bm{v}^e$ and the standard deviation predicts the uncertainty in the action.
Compensating for the nonlinearity of $\bm{v}^e$ is the basis for our \emph{nonlinear mismatch controller}, which we present next.
A control block diagram of our control approach is shown in Fig. \ref{fig:control-diagram}.

\subsection{Optimization-Based Controller}

To construct our optimization-based controller, we start by replacing the desired velocity $\bm{v}^d$ in our error dynamics \eqref{eq:vError} with our model.
This yields the actual velocity of the $\mu$bot for any given heading angle $\alpha$ in the form:
\begin{equation} \label{eq:v=vd}
    \hat{a}_0\,f\begin{bmatrix}
        \cos(\alpha) \\ \sin(\alpha)
    \end{bmatrix} + \hat{\bm{D}} + \bm{v}^e(\alpha) = \bm{v}.
\end{equation}
Next, we replace the unknown velocity error $\bm{v}^e$ with the mean prediction from the GP.
This assumes that the GP has sufficiently learned the velocity error function $\bm{v}_e$, which yields
\begin{equation} \label{eq:alphaFunc}
    \hat{a}_0 \, f \, \begin{bmatrix}
        \cos(\alpha) \\ \sin(\alpha)
    \end{bmatrix}
    + \bm{\hat{D}}
    + \bm{\mu}(\alpha) = \bm{v}(\alpha),
\end{equation}
where $\bm{\mu}$ is the GP's estimate for each component of the velocity error.
To minimize the velocity error of our $\mu$bot, we perform a least squares minimization of \eqref{eq:alphaFunc} from the desired velocity, i.e.,
\begin{equation} \label{eq:lsq}
    \min_{\alpha} ||\bm{v}(\alpha) - \bm{v}^d||^2.
\end{equation}
Applying Assumptions \ref{smp:disturbance}--\ref{smp:constantSpeed} and expanding \eqref{eq:lsq} yields a one-dimensional least-squares cost function
\begin{align}
    J(\alpha) =& \Big(\hat{a}_0 f \cos(\alpha) + \mu_x(\alpha) + \hat{D}_x - v^d_x \Big)^2 \notag\\
    &+ \Big(\hat{a}_0 f \sin(\alpha) + \mu_y(\alpha) + \hat{D}_y - v^d_y \Big)^2.
\end{align}
Expanding the cost function and applying the Pythagorean identity yields the following least-squares optimization problem
\begin{align} \label{eq:optimization}
    \min_{\alpha\in[-\pi,\pi]} &(\hat{a}_0 f)^2 + ||\bm{\mu}(\alpha) + \hat{\bm{D}} - \bm{v}^d||^2 \notag\\
    &+ 2 \hat{a}_0 f \Big(\bm{\mu}(\alpha) + \hat{\bm{D}} - \bm{v}^d\Big) \cdot
    \begin{bmatrix}
    \cos(\alpha) \\ \sin(\alpha)
    \end{bmatrix},
\end{align}
which is a differentiable scalar optimization problem over a compact set.

\subsection{Online Learning} \label{sec:howto-learn}

We train the GP during an initial \emph{learning phase}, where the $\mu$bot is given a sequence of control inputs, either from a human operator or open-loop control sequence.
We collect position and control action data for the $\mu$bot at discrete time steps $t_k$; we denote the position data by $\mathcal{P} = \big\{ \bm{p}(t_k) \big\}$ and action data as $\mathcal{X} = \big\{\alpha(t_k) \big\}$.
We calculate the actual velocity $\bm{v}(t_k)$ by taking a numerical derivative of $\mathcal{P}$ and passing the result through a low-pass filter; this yields the actual velocity of the $\mu$bot at each step, which we store in the set $\mathcal{V} = \{ \bm{v}(t_k) \}$. 

Once the data is collected, we estimate the model parameters and desired velocity as follows. First, we estimate $\bm{D}$ by applying a control input of $f(t) = \alpha(t) = 0$, which yields
\begin{equation}
    \dot{\bm{p}} = \bm{D}.
\end{equation}
Taking the expectation of both sides yields the mean disturbance
\begin{equation} \label{eq:estimateD}
    \frac{1}{|\mathcal{V}|}\sum_{\bm{v}(t_k)\in\mathcal{V}} ||\bm{v}(t_k)|| = \mathds{E}\big[\bm{D}\big] = \hat{\bm{D}},
\end{equation}
where $|\cdot|$ is set cardinality.

Next, we determine $\hat{a}_0$ using data from an open loop control sequence; taking the expectation of \eqref{eq:dynamics} and squaring both sides yields
\begin{equation}
    \mathds{E}\big[ ||\bm{v} - \hat{\bm{D}}||\big]^2 = a_0^2 f^2.
\end{equation}
Substituting the expectation of $\bm{v}$ with the experimental data, re-arranging, and taking the square root of both sides yields the best statistical estimate for $\hat{a}_0$:
\begin{equation} \label{eq:estimateA}
    \hat{a}_0 = \frac{1}{|\mathcal{V}|}\sum_{\bm{v}(t_k)\in\mathcal{V}}\frac{||\bm{v}(t_k) - \hat{\bm{D}}||}{f}.
\end{equation}

Finally, the resulting set of velocity errors is
\begin{equation} \label{eq:vErrorSet}
    \mathcal{Y} = \Big\{ \bm{v}^e(t_k) ~:~ \bm{v}^e = \bm{v}(t_k) - \bm{v}^d(t_k) \Big\},
\end{equation}
where $\bm{v}^{d}(t_k)$ is the desired velocity of the $\mu$bot at each time $t_k$. We use \eqref{eq:vErrorSet} in conjunction with our empirical model \eqref{eq:model-flatness} to generate the velocity error and control action at each time step.
With this data, we compute a posterior distribution on the mean and standard deviation of the GP to determine the expected velocity error and its uncertainty for each control input.

\section{Experimental Results} \label{sec:results}

We validated our learning approach {\it in silico} and {\it in situ}\footnote{Videos of the experiments and supplemental material are available online: \url{https://sites.google.com/udel.edu/l4ub} }; we present our simulation results in Subsection \ref{sec:sim} and experimental findings in Subsection \ref{sec:exp}.
In both cases, we first perform an online learning step, where we apply a pre-computed control input to generate training data.
Then, to validate our learning approach, we apply a pre-defined sequence of control actions in open-loop with and without the Nonlinear Mismatch module (Fig. \ref{fig:control-diagram}).
This yields the \emph{corrected} and \emph{baseline} cases, respectively, which we use to explicitly quantify the impact of our approach.

We implemented our GP approach using the Scikit-Learn toolbox (see \cite{scikit-learn}) for Python3, which provides an API to easily select a large number of kernels and train the GP.
Scikit-learn also automatically optimizes the kernel hyperparameters during training, which provided insights for kernel selection.
In particular, some hyperparameters for the rational quadratic, Matern, and periodic kernels grew arbitrarily small during training, which implies that these kernels include extraneous dynamics that do not describe the true behavior of the rolling $\mu$bot's velocity error.
We found that a linear combination of a radial basis function and white noise in the form 
\begin{equation}
    K(\alpha, \alpha') = \exp{\frac{||\alpha - \alpha'||^2}{2\sigma}} + \eta
\end{equation}
yielded a kernel that adequately captured the velocity error of the rolling $\mu$bot. In the equation above, $\sigma$ is a length hyperparameter and $\eta$ is drawn from a normal distribution where the mean is zero and the variance is another hyperparameter.

\subsection{In Silico Experiment}\label{sec:sim}

We developed a $\mu$bot simulator as an \emph{OpenAI Gym}\footnote{For more information on the Gym environment see: \url{https://github.com/openai/gym}} environment in Python3.
We implemented two simulation modes using a `\emph{model-mismatch}' flag, which disturbs the model parameters and adds stochastic zero-mean noise to mimic a physical experiment.
Omitting this flag uses the exact model parameters with no noise to generate the desired system trajectory.
We implemented the learning approach of Section \ref{sec:learning} as follows. First, we applied zero input over $100$ time steps ($3$ seconds) with the `model-mismatch' flag to estimate the mean disturbance using \eqref{eq:estimateD}.
Next, we generated a sequence of control inputs that swept the entire control domain $\alpha\in[-\pi, \pi]$ three times over $1800$ time steps ($60$ seconds) with the `model-mismatch' flag, which produced our training data.
In training, we estimated $\hat{a}_0$ and updated the GPs using \eqref{eq:estimateA} and \eqref{eq:vErrorSet}, respectively.
Finally, to validate our approach, we performed three experiments in silico; 1) we generated the desired trajectory without the `model-mismatch' flag, 2)
we generated the \emph{baseline} trajectory by repeating the experiment with the `model-mismatch' flag enabled, and 3) we updated the reference control inputs using \eqref{eq:optimization} to generate the \emph{corrected} trajectory with the `model mismatch' flag.

\begin{figure}[ht]
    \centering
    \includegraphics[width=0.8\linewidth]{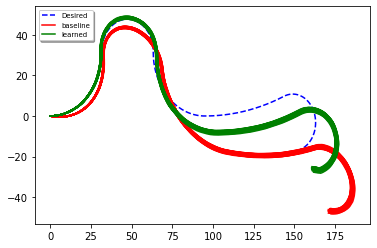}
    \caption{The desired, (blue, dashed), corrected (green), and baseline (orange) trajectories from 100 different trials of the {\it in silico} experiment.}
    \label{fig:simulation}
\end{figure}

\begin{figure}[ht]
    \centering
    \includegraphics[width=0.8\linewidth]{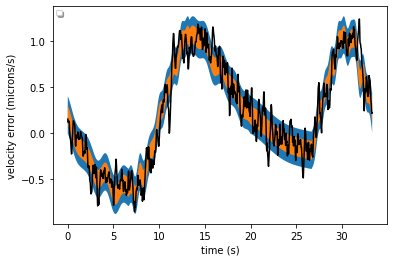}
    \caption{The GP's prediction of the $x$-axis velocity errors at each time step; the orange band corresponds to one standard deviation (65\%), and the blue band corresponds to two standard deviations (95\%) of uncertainty. The black line is the actual velocity error.}
    \label{fig:error-estimate-vx}
\end{figure}

\begin{figure}[ht]
    \centering
    \includegraphics[width=0.8\linewidth]{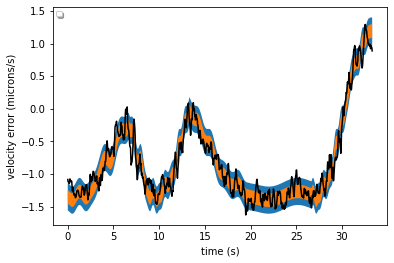}
    \caption{The GP's prediction of the $y$-axis velocity errors at each time step; orange band corresponds to one standard deviation (65\%), and the blue band corresponds to two standard deviations (95\%) of uncertainty. The black line is the actual velocity error.}
    \label{fig:error-estimate-vy}
\end{figure}

Fig. \ref{fig:simulation} shows the resulting desired, baseline, and corrected trajectories overlaid for 100 trials with the same initial state.
While the learning component significantly improves the velocity tracking, it is unable to completely compensate for the model mismatch--even in an environment with no noise.
The velocity error estimate for one trial is presented in Figs. \ref{fig:error-estimate-vx} and \ref{fig:error-estimate-vy}, which demonstrates that the GP has captured a reasonably good estimate of the velocity error at each time step.
This implies that the nonlinear mismatch approach is unable to achieve perfect tracking for the system, which is likely related to the reachability of the system's dynamics \eqref{eq:alphaFunc}.
This stems from correcting the $x$ and $y$ components of the velocity error while only controlling $\alpha$.

\subsection{In Situ Experiment} \label{sec:exp}

\begin{figure}[ht]
\vspace{1em}
    \centering
    \includegraphics[width=0.8\linewidth]{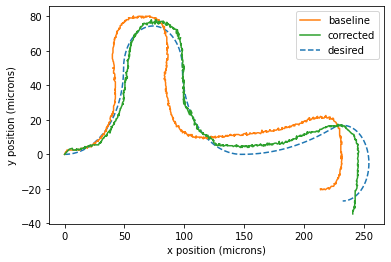}
    \caption{A comparison of the baseline (orange), corrected (green), and desired (blue, dashed) trajectories from the {\it in situ} experiment.}
    \label{fig:experiment-traj}
\end{figure}

\begin{figure}[ht]
    \centering
    \includegraphics[width=0.8\linewidth]{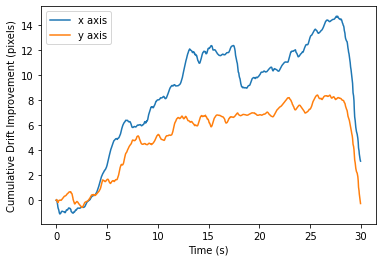}
    \caption{Improvement in the cumulative drift of the $\mu$bot between the baseline and corrected cases for the {\it in situ} experiment.}
    \label{fig:experiment-drift}
\end{figure}

\begin{figure*}[ht]
\vspace{1em}
    \centering
    \includegraphics[width=0.9\textwidth]{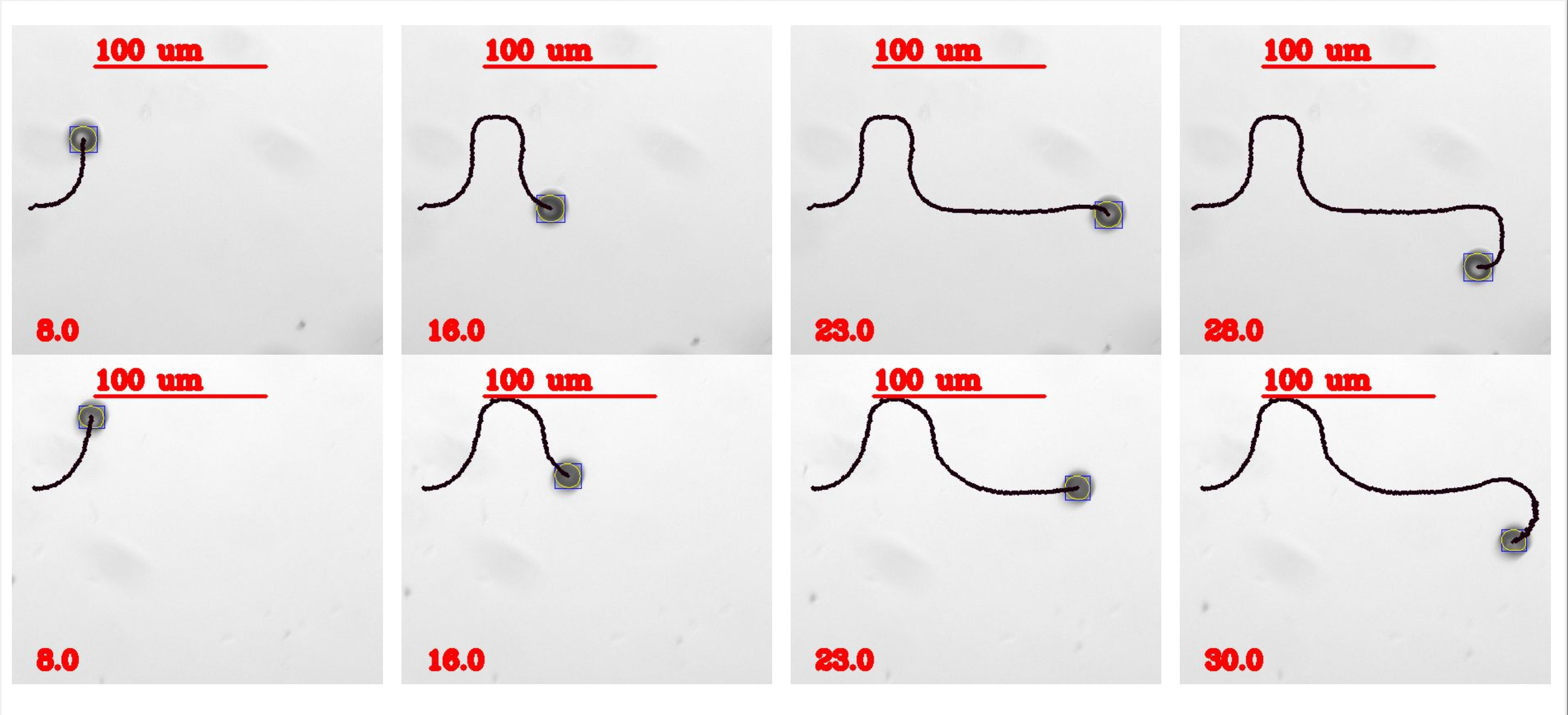}
    \caption{Snapshots of a different set of \emph{baseline} (top) and \emph{corrected} (bottom) experiments, taken approximately 8 seconds apart and showing the $\mu$bot with the position history overlaid.}
    \label{fig:snapshot}
\end{figure*}

We repeated the same procedure from Section~\ref{sec:sim} using $24$um $\mu$bots over $300$ time steps ($10$ seconds) at the experimental facility at the University of Delaware as described in Section \ref{sec:experimentSetup}.
The resulting desired, baseline, and corrected trajectories are presented in Fig. \ref{fig:experiment-traj}, and the drift error for the baseline and corrected cases is shown in Fig. \ref{fig:experiment-drift}.
Photos of an experiment with the trajectories overlaid are presented in Fig. \ref{fig:snapshot}.

Fig. \ref{fig:experiment-traj} shows significant improvement in the $\mu$bot's ability to track the open-loop trajectory.
We translated the normalized and corrected position data to the origin to compare it with the desired traejctory.
As a result, the clear improvement in the $\mu$bot's position trajectory tracking comes from a combination of our improved tracking controller and random disturbances.
In other words, integrating the $\mu$bot's velocity using our corrected control signal yields less error than the uncorrected case.
This is shown explicitly in Fig. \ref{fig:experiment-drift}, which depicts the $\mu$bot's drift throughout the experiment.
To calculate the $\mu$bot's drift, we subtracted the desired and actual velocity along each axis to calculate the velocity error.
Next, we performed a cumulative trapezoidal integration on the absolute value of the velocity error, which quantified the worst-case scenario for how far the $\mu$bot could drift from the reference trajectory.
As a result, drift was reduced by at least $6$ pixels ($3.7$ microns) along each axis for the majority of the experiment.

The median velocity error along each axis is presented in Table \ref{tab:experiment}, along with the error in the $\mu$bot's final position for each case.
These results show that despite the poor tracking in the final $3$ seconds, our learning controller significantly reduces the drift of the $\mu$bot by matching the desired open-loop control policy and brings the $\mu$bot closer to the desired final position.
Due to the nature of the experimental environment, it is not uncommon for unexpected disturbances, such as stiction, debris, and nearby magnetic particles, to disturb the $\mu$bot's trajectory in a way that our tracking controller cannot compensate for.
These exogenous factors are the source of error in the last $3$ seconds of the \emph{corrected} experiment.

\begin{table}[ht]
    \centering
    \begin{tabular}{c|clclc}
                              & Baseline & Corrected & Improvement \\ \toprule
         Final Position Error & 12.56 microns & 7.14 microns  & 43 \% \\ 
         Median $v_x$ Error   & 1.19 microns/s & 0.72 microns/s & 40 \% \\ 
         Median $v_y$ Error   & 0.80 microns/s & 0.61 microns/s & 23 \% 
    \end{tabular}
    \caption{Errors for the {\it in situ} $\mu$bot experiment; the median uses the absolute value of the error.}
    \label{tab:experiment}
\end{table}

\section{Conclusion} \label{sec:conclusions}

We developed a nonlinear mismatch controller to improve the performance of a tracking controller in $2D$.
We motivated the use of nonlinear mismatch over a parameter estimation scheme, and we proposed a least-squares based optimization problem to minimize the tracking error.
Finally, we demonstrated in simulation and experiments that our approach significantly improves the tracking performance of rolling $\mu$bots.

Future work includes relaxing Assumption \ref{smp:constantSpeed} and including $f$ as parameter in the model mismatch.
Deriving guarantees on the resulting velocity error using fixed-point analysis is another interesting research direction; employing the GP's uncertainty estimate as a measure of robustness in a high-level planner may also yield useful insights.
Embedding our low-level controller inside of an MPC path planner to avoid undesired collisions with cells and counteract Brownian diffusion in-situ is another critical next step for this work.
Finally, expanding our approach to control multiple $\mu$bots simultaneously would advance the state of the art, and bring us one step closer to solving fundamental challenges in emerging medical applications.

\bibliography{mendeley,bibliography,microbots}

\begin{thebibliography}{10}

\bibitem{honda1996micro}
T.~Honda, K.~Arai, and K.~Ishiyama, ``Micro swimming mechanisms propelled by
  external magnetic fields,'' {\em IEEE Transactions on Magnetics}, vol.~32,
  no.~5, pp.~5085--5087, 1996.

\bibitem{sitti2015biomedical}
M.~Sitti, H.~Ceylan, W.~Hu, J.~Giltinan, M.~Turan, S.~Yim, and E.~Diller,
  ``Biomedical applications of untethered mobile milli/microrobots,'' {\em
  Proceedings of the IEEE}, vol.~103, no.~2, pp.~205--224, 2015.

\bibitem{troccaz2008development}
J.~Troccaz and R.~Bogue, ``The development of medical microrobots: a review of
  progress,'' {\em Industrial Robot: An International Journal}, 2008.

\bibitem{Barcena2009ApplicationsBiomedicine}
C.~B{\'a}rcena, A.~K. Sra, and J.~Gao, ``{Applications of magnetic
  nanoparticles in biomedicine},'' in {\em Nanoscale Magnetic Materials and
  Applications}, 2009.

\bibitem{Guo2007MechanismApplication}
S.~Guo and Q.~Pan, ``{Mechanism and control of a novel type microrobot for
  biomedical application},'' in {\em Proceedings 2007 IEEE International
  Conference on Robotics and Automation}, pp.~187--192, 2007.

\bibitem{Sakar2011WirelessMicrotransporters}
M.~S. Sakar, E.~B. Steager, A.~Cowley, V.~Kumar, and G.~J. Pappas, ``{Wireless
  manipulation of single cells using magnetic microtransporters},'' in {\em
  2011 IEEE International Conference on Robotics and Automation},
  pp.~2668--2673, 2011.

\bibitem{Jagerpaper}
E.~W.~H. Jager, O.~Ingan{\"{a}}s, and I.~S. Lundstr{\"{o}}m, ``{Microrobots for
  micrometer-size objects in aqueous media: potential tools for single-cell
  manipulation},'' vol.~288, no.~5475, pp.~2335--2338, 2000.

\bibitem{kim2013fabrication}
S.~Kim, F.~Qiu, S.~Kim, A.~Ghanbari, C.~Moon, L.~Zhang, B.~J. Nelson, and
  H.~Choi, ``Fabrication and characterization of magnetic microrobots for
  three-dimensional cell culture and targeted transportation,'' {\em Advanced
  Materials}, vol.~25, no.~41, pp.~5863--5868, 2013.

\bibitem{Steager2013AutomatedMicrorobots}
E.~B. Steager, M.~Selman~Sakar, C.~Magee, M.~Kennedy, A.~Cowley, and V.~Kumar,
  ``{Automated biomanipulation of single cells using magnetic microrobots},''
  vol.~32, no.~3, pp.~346--359, 2013.

\bibitem{kim2015electric}
H.~Kim and M.~J. Kim, ``Electric field control of bacteria-powered microrobots
  using a static obstacle avoidance algorithm,'' {\em IEEE Transactions on
  Robotics}, vol.~32, no.~1, pp.~125--137, 2015.

\bibitem{palima2013gearing}
D.~Palima and J.~Gl{\"u}ckstad, ``Gearing up for optical microrobotics:
  micromanipulation and actuation of synthetic microstructures by optical
  forces,'' {\em Laser \& Photonics Reviews}, vol.~7, no.~4, pp.~478--494,
  2013.

\bibitem{chowdhury2016towards}
S.~Chowdhury, W.~Jing, and D.~J. Cappelleri, ``Towards independent control of
  multiple magnetic mobile microrobots,'' {\em Micromachines}, vol.~7, no.~1,
  p.~3, 2016.

\bibitem{ErdemThermallyMicrorobot}
E.~Y. Erdem, Y.-M. Chen, M.~Mohebbi, J.~W. Suh, G.~T. Kovacs, R.~B. Darling,
  and K.~F. B{\"o}hringer, ``Thermally actuated omnidirectional walking
  microrobot,'' {\em Journal of Microelectromechanical Systems}, vol.~19,
  no.~3, pp.~433--442, 2010.

\bibitem{Behkam2007BacteriaMicrorobots}
B.~Behkam and M.~Sitti, ``{Bacteria integrated swimming microrobots},'' in {\em
  Lecture Notes in Computer Science}, vol.~4850 LNAI, pp.~154--163, 2007.

\bibitem{pieters2015rodbot}
R.~Pieters, H.-W. Tung, S.~Charreyron, D.~F. Sargent, and B.~J. Nelson,
  ``Rodbot: A rolling microrobot for micromanipulation,'' in {\em 2015 IEEE
  International Conference on Robotics and Automation (ICRA)}, pp.~4042--4047,
  2015.

\bibitem{Rivas2022}
D.~Rivas, S.~Mallick, M.~Sokolich, and S.~Das, ``Cellular manipulation using
  rolling microrobots,'' in {\em 2022 International Conference on Manipulation,
  Automation and Robotics at Small Scales (MARSS)}, pp.~1--6, 2022.

\bibitem{das2018experiments}
S.~Das, E.~B. Steager, M.~A. Hsieh, K.~J. Stebe, and V.~Kumar, ``Experiments
  and open-loop control of multiple catalytic microrobots,'' {\em Journal of
  Micro-Bio Robotics}, vol.~14, no.~1-2, pp.~25--34, 2018.

\bibitem{vinagre2016there}
B.~M. Vinagre, I.~Tejado, and J.~E. Traver, ``There's plenty of fractional at
  the bottom, i: Brownian motors and swimming microrobots,'' {\em Fractional
  Calculus and Applied Analysis}, vol.~19, no.~5, p.~1282, 2016.

\bibitem{xu2015magnetic}
T.~Xu, J.~Yu, X.~Yan, H.~Choi, and L.~Zhang, ``Magnetic actuation based motion
  control for microrobots: An overview,'' {\em Micromachines}, vol.~6, no.~9,
  pp.~1346--1364, 2015.

\bibitem{tang2022vision}
X.~Tang, Y.~Li, X.~Liu, D.~Liu, Z.~Chen, and T.~Arai, ``Vision-based automated
  control of magnetic microrobots,'' {\em Micromachines}, vol.~13, no.~2,
  p.~337, 2022.

\bibitem{behrens2022smart}
M.~R. Behrens and W.~C. Ruder, ``Smart magnetic microrobots learn to swim with
  deep reinforcement learning,'' {\em Advanced Intelligent Systems}, vol.~4,
  no.~10, p.~2200023, 2022.

\bibitem{Greeff2021ExploitingProcesses}
M.~Greeff and A.~P. Schoellig, ``{Exploiting Differential Flatness for Robust
  Learning-Based Tracking Control using Gaussian Processes},'' {\em IEEE
  Control System Letters}, vol.~5, no.~4, pp.~1121--1126, 2021.

\bibitem{Yang2020}
L.~Yang, Y.~Zhang, Q.~Wang, K.-F. Chan, and L.~Zhang, ``Automated control of
  magnetic spore-based microrobot using fluorescence imaging for targeted
  delivery with cellular resolution,'' {\em IEEE Transactions on Automation
  Science and Engineering}, vol.~17, no.~1, pp.~490--501, 2020.

\bibitem{Dongsik2017MotionVehicles}
D.~Chang, W.~Wu, C.~R. Edwards, and F.~Zhang, ``Motion tomography: Mapping flow
  fields using autonomous underwater vehicles,'' {\em The International Journal
  of Robotics Research}, vol.~36, no.~3, pp.~320--336, 2017.

\bibitem{Beaver2021AMicrorobots}
L.~E. Beaver, B.~Wu, S.~Das, and A.~A. Malikopoulos, ``A first-order approach
  to model simultaneous control of multiple microrobots,'' in {\em 2022
  International Conference on Manipulation, Automation and Robotics at Small
  Scales (MARSS)}, pp.~1--7, 2022.

\bibitem{Sira-Ramirez2018DifferentiallySystems}
H.~Sira-Ramirez and S.~K. Agrawal, {\em {Differentially Flat Systems}}.
\newblock 1st~ed., 2018.

\bibitem{scikit-learn}
F.~Pedregosa, G.~Varoquaux, A.~Gramfort, V.~Michel, B.~Thirion, O.~Grisel,
  M.~Blondel, P.~Prettenhofer, R.~Weiss, V.~Dubourg, J.~Vanderplas, A.~Passos,
  D.~Cournapeau, M.~Brucher, M.~Perrot, and E.~Duchesnay, ``Scikit-learn:
  Machine learning in {P}ython,'' {\em Journal of Machine Learning Research},
  vol.~12, pp.~2825--2830, 2011.

\end{thebibliography}

\end{document}